\documentclass[conference]{IEEEtran}
\usepackage{graphicx}
\usepackage{amsmath}	
\usepackage{cite}
\usepackage{footnote}
\makesavenoteenv{tabular}
\IEEEoverridecommandlockouts

\usepackage{tablefootnote}
\usepackage{url}
\begin{document}

\title{Learning sentence embeddings using Recursive Networks}
\author
{\IEEEauthorblockN{Anson Bastos} 
\IEEEauthorblockA{Independent scholar\\
Mumbai, India \\
ansonbastos@gmail.com\\
}
}

\maketitle

\begin{abstract}
Learning sentence vectors that generalise well is a challenging task. In this paper we compare three methods of learning phrase embeddings: 1) Using LSTMs, 2) using recursive nets, 3) A variant of the method 2 using the POS information of the phrase. We train our models on dictionary definitions of words to obtain a reverse dictionary application similar to Felix et al. \cite{2}. To see if our embeddings can be transferred to a new task we also train and test on the rotten tomatoes dataset \cite{1}. We train keeping the sentence embeddings fixed as well as with fine tuning.
\end{abstract}

\section{Introduction}
Many methods to obtain sentence embeddings have been researched in the past. Some of these include Hamid at al. \cite{5}. who have used LSTMs and Ozan et al. \cite{6} who have made use of the recursive networks. Another approach used by Mikolov et al.  \cite{6} is to include the paragraph in the context of the target word and then use gradient descent at inference time to find a new paragraph’s embedding. This method suffers from the obvious limitation that the number of paragraphs could grow exponentially with the number of words. Chen \cite{7} has used the average of the embeddings of words sampled from the document. Lin et al. \cite{8} have used a bidirectional LSTM to find the sentence embeddings which is further used for classification tasks.\\
There have been attempts in the past to train sentence vectors using dictionary definitions \cite{2}. Tissier et al. \cite{3} have trained word embeddings similar to Mikolov et al. \cite{4} using dictionary definitions. However they have not used these embeddings in the reverse dictionary application.\\
We train the LSTM model on an enlarged unsupervised set of data where we randomise the words in the sentence as seen in section 4.\\
The results of our experiments are as follows:\\
1)  Randomising the sentences increases the accuracy of the LSTM model by 25-40\%\\
2) RNNs perform better than LSTM given limited data. The top 3 test set accuracy on the reverse definition task increases by 40\% using RNN as compared to LSTM on the same ammount of data
3)  Pretraining the LSTM model with the dictionary and fine-tuning does not change the classification accuracy on the test set but it becomes difficult to overfit the training set\\
All code is made publicly available. \footnote{\url{https://github.com/ansonb/reverse_dictionary}}

\section{System Architecture}
We have made use of three architectures as described below\\

\subsection{LSTM}
We make use of the basic LSTM architecture as proposed by Zaremba et al. \cite{9}. The equations describing the network is as follows:\\
\begin{equation}
i = \sigma(W_{i_{1}}*h_{t}^{l-1} + W_{i_{2}}*h_{t-1}^{l} + B_{i})
\end{equation}
\begin{equation}
f = \sigma(W_{f_{1}}*h_{t}^{l-1} + W_{f_{2}}*h_{t-1}^{l} + B_{f}\\
\end{equation}
\begin{equation}
o = \sigma(W_{o_{1}}*h_{t}^{l-1} + W_{g_{2}}*h_{t-1}^{l} + B_{g})\\
\end{equation}
\begin{equation}
g = tanh(W_{g_{1}}*h_{t}^{l-1} + W_{o}^{2}*h_{t-1}^{l} + B_{g})\\
\end{equation}
\begin{equation}
c_{t}^{l} = f{\odot}c_{t-1}^{l} + i{\odot}g\\
\end{equation}
\begin{equation}
h_{t}^{l} = o{\odot}tanh(c_{t}^{l})\\
\end{equation}

where $h_{t}^{l-1}$ is the output of the previous layer or the input\\
$h_{t-1}^{l}$ is the output of the previous time step of the same layer

\begin{figure}[h]
\centering
\includegraphics[scale =0.5]{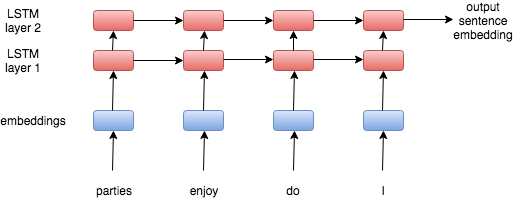}
\caption{The architecture of the LSTM model. The example sentence used is 'I do enjoy parties'}
\end{figure}

The words are fed to the network in the reverse order as suggested by Sutskever et al. \cite{11}. The embeddings from the embedding matrix are then fed to a uni directional LSTM containing 2 layers and 256 hidden units each. We make use of the output of the last layer at the final timestep as the vector representation of the sentence.\\

\subsection{Recursive neural network (RNN) with shared weights}
In the recursive network we form the parse tree using the syntaxnet parser \cite{10}. For any node in the parsed tree the following equation is computed\\
\begin{equation}
f(i) = relu(E_{i}*W + b + f_{children}(i))
\end{equation}
\begin{equation}
f_{children}(i) = \sum_{j\ {\in{\ children\ of\ node\ i}}}(f(j))
\end{equation}

where $E_{i}$ is the embedding of the word at the ith node\\ 

\begin{figure}[h]
\includegraphics[width=0.5\textwidth]{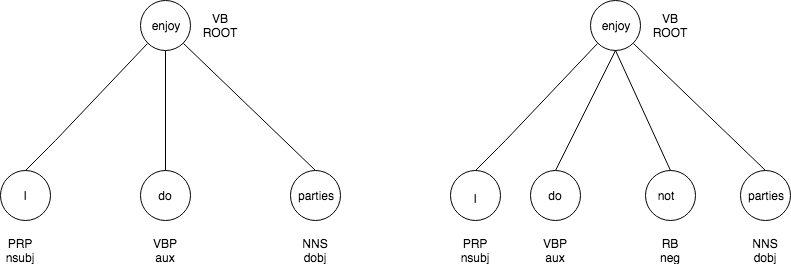}
\caption{The parse tree structure of the sentences 'I do enjoy parties' and 'I do not enjoy parties' obtained using syntaxnet}
\end{figure}

\begin{figure}[h]
\includegraphics[width=0.5\textwidth]{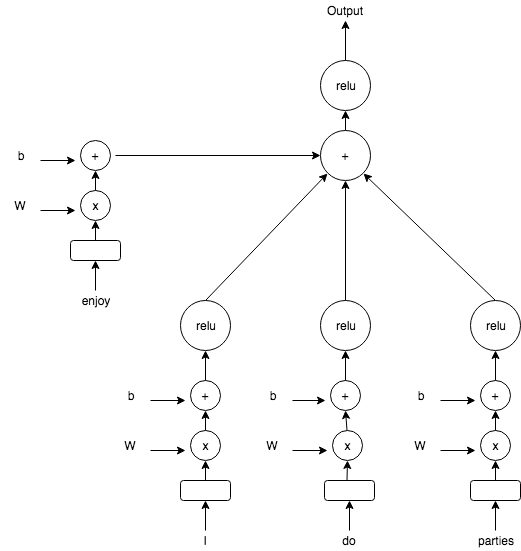}
\caption{The architecture of the RNN model shared weights}
\end{figure}

Here we make use of a common shared matrix W.\\

\subsection{RNN with unique weights}
This method is similar to the above method except that the weights W are different for different POS words and the output of each node is multiplied by a weight .\\

\begin{equation}
f(i) = relu(E_{i}*W(i) + b + f_children(i))
\end{equation}

\begin{equation}
f_children(i) = \sum_{j{\in{children of node i}}}( f(j)*w(j) )
\end{equation}

$w(j)$ is the weightage between -1 and 1 to be given to the output at node j and is decided by the word at node j and its immediate children.\\

Algorithm 1: Find w at node i\\
=============================\\
1) Initialise an array arr as empty\\
2) Fill the current node and its children in a nodes in arr\\
3) for j in nodes :\\
4) \hspace{1cm} arr.append(classifier(E(j)))\\
5) $w(i) = tanh(maxpool(arr))$ \\
6) return $w(i)$

\begin{figure}[h]
\includegraphics[width=0.5\textwidth]{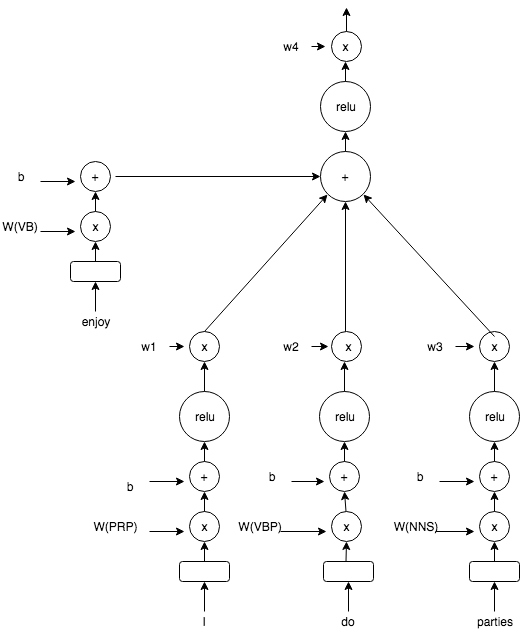}
\caption{ The architecture of RNN with separate weights for each POS and taken a weighted sum}
\end{figure}

\begin{figure}[h]
\includegraphics[width=0.5\textwidth]{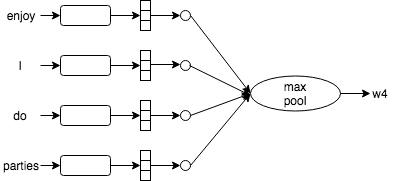}
\caption{ The architecture for determining the weight at node enjoy}
\end{figure}

The intuition behind this approach is that not all words will have the same processing. For example consider the sentence “I do enjoy parties” vs the sentence “I do not enjoy parties”. In both the only difference is the negative word not. Let’s consider that the model learns to take the average of all the word embeddings. In this case it will not be able to properly model each negative case of every sentence. Ideally we would expect the embedding of a negative sentence to be in the opposite direction of the corresponding positive sentence. This model consists of the weighted sum of all the embeddings, the weights being decided by the current node and its direct children nodes if any. So in case of the sentence “I do not enjoy parties” the final embedding at root enjoy should be multiplied by a negative weight to give the opposite of “I do enjoy parties”. Another advantage of this method is that it could lean to give more importance to some sibling nodes and lesser importance to others, as in the case of conjunctions. The algorithm of the method to find the weights at each node is given above in algorithm 1.

\section{Objective function}
The sentence embedding from all of the methods above is multiplied with the word embeddings to find the closest matching word from the dictionary. \\
Let Es be the sentence embedding, Ed be the embeddings of the dictionary words. then the output word is given as\\

\begin{equation}
logits = E_{s}^{T}*E_{d}
\end{equation}

\begin{equation}
Output\ word = argmax(logits)
\end{equation}

In all the above three methods we minimise the cross entropy loss of the output.\\

\begin{equation}
\resizebox{.45 \textwidth}{!}
{
\textit{loss = labels*-log($\sigma$(logits)) + (1-labels)*-log(1-$\sigma$(logits))}
}
\end{equation}

According to the above equation the first term will cause similar sentences to move closer together and the second term will cause the dissimilar sentence embeddings to move away from each other

\section{Preparing the data}
The Webster’s dictionary provided by the Gutenberg project was was used for obtaining the word definitions. After processing the dictionary 296,39 definitions were obtained. The dataset contains definitions of 95,831 unique words. The vocabulary size was around 138,443.\\
As training the entire set of words takes around a month on a dual core system with 4 GB RAM we have taken a small subset of 144 words only for comparison between the three models. The test set was prepared by manually framing the definitions for these words. For training with the LSTM we randomise the sentences to obtain 10, 100 and 1000 times the data. We compare the performance on the test set for these three augmented datasets using only the LSTM as it would not make any sense to obtain a parse tree from randomised sentences. \\
For the classification task we take all the definitions of the words present in the rotten tomatoes training dataset. We obtain 7779 unique word definitions and a vocabulary size of 40,990.

\section{Experimental setup and training}
The word embeddings are of size 32. For a large vocabulary Felix et al. \cite{2} suggest using a size of 200-500. However for our sample dataset a dimension of 32 suffices. For the LSTM we have made use of 2 layers each containing 256 hidden units. For the RNNs the Weight matrix is of size 32x32. The hidden layer used for finding the weights for the weighted sum uses 10 units.\\
The training was done on a dual core and 4 GB RAM ubuntu system. It takes around 3-4 hours to train the LSTM on the 144 words augmented (144*1000) dataset, 4 hours to train the RNNs on the 144 words dataset, 5-6 hours for training the LSTM on the rotten tomatoes vocabulary.
For RNNs we split the data into buckets of maximum tree levels, of 2, 4, 6 and so on till 14, in each bucket. This helps in making the tree size uniform for all examples in each bucket for training. Having a single uniform tree size would consume a huge amount of memory, around 3 million nodes or 63 MB per training data. The extra nodes in each example are filled with -1s and their embeddings are kept a constant of value 0.

\section{Results}
\begin{table}[!htbp]
\centering
\label{tab :result}
\begin{tabular}{|c|p{1cm}|p{1cm}|p{1cm}|p{1cm}|}
\hline
Method & Wout seperate & Data Augmentation & Accuracy (top 1) & Accuracy (top 3)\\
\hline
LSTM & Yes & x1 & 24.32\% & 29.73\%\\
\hline
LSTM & No & x1 & 27.03\% & 29.73\%\\
\hline
RNN with shared weights & Yes & x1 & \bf{54.05}\% & \bf{72.97}\%\\
\hline
RNN with shared weights & No & x1 & 45.94\% & 64.86\%\\
\hline
RNN with separate weights & Yes & x1 & 43.24\% & 62.16\%\\
\hline
RNN with separate weights & No & x1 & 27.03\% & 37.84\%\\
\hline
LSTM & No & x1 & 21.62\% & 27.02\%\\
\hline
LSTM & No & x1 & 54.05\% & 67.57\%\\
\hline
LSTM & No & x1 & 51.35\% & 70.27\%\\
\hline
\end{tabular}
\caption{Comparison of LSTMs with RNNs on the sampled words dataset. In this we present the accuracy of the three methods trained on the dataset of 144 words and tested on 37 words.}
\end{table}
 
It can be seen from table 1 that RNNs have a higher accuracy as compared to the LSTMs when trained with same data. Even with 1000 times more data the LSTMs perform only comparably but not very better. It may be because of the padding used in the LSTMs, for making all the sequences of a constant maximum size of 66 in our case. Another reason could be that the LSTMs use much higher dimension (256 vs 32 for RNN) and so require more data to prevent overfitting.\\
We try two combinations of the output embedding keeping it same as the input embedding and a more general case of a new embedding. The new embedding may or may not learn the same embeddings as the original input. We observe that the accuracy increases when a separate embedding is used. It seems that it is more difficult to obtain an embedding for the sentence in the word embedding space.\\
 
 \begin{table}[!htbp]
 \centering
 \label{tab :result}
 \begin{tabular}{|p{2cm}|p{1cm}|p{1cm}|}
 \hline
 Method & Training set accuracy & Test set accuracy\\
 \hline
 LSTM end to end training & 96\% & 60\%\\
 \hline
 LSTM pertained on dictionary words without fine tuning & 50\% & 43\%\\
 \hline
 LSTM pertained on dictionary words with fine tuning & 70\% & 60\%\\
 \hline
 \end{tabular}
 \caption{Comparison of accuracies obtained on the rotten tomatoes dataset. We compare the training and test accuracies of the three methods using only LSTMs.}
 \end{table}
 
In the table 2 we see that there is a huge difference in the training and test set accuracy when we train end to end using LSTMs without pre training on the dictionary words. When we retrain using the LSTM and keep the embeddings fixed and train only the classification layer the training accuracy falls to 50\% and so does the test accuracy. On pre training with the dictionary words and then fine-tuning the embeddings the test set accuracy is maintained and the training accuracy is also closer to the test accuracy. We were expecting an increase in the test accuracy without a change in the training accuracy. However the results suggest that overfitting becomes difficult when pre trained. Every method has been trained for around 50,000 steps. When the pre trained model was trained further to 200,000 steps the training accuracy increases to only 80\% and the test accuracy decreases to 57\%. Further research might help in confirming that fine tuning with such a pretrained model could give a correct representation of the actual error in the absence of a test set.\\
 
\section{Conclusion}
n this paper we try to implement a different method to find sentence embeddings using recursive neural networks (RNN). We find out that RNNs perform at least comparably to LSTMs at a much lesser parameters (32 compared to 256). We train word embeddings using the LSTM and use these pretrained embeddings on the rotten tomatoes dataset and find out that pertaining makes it difficult to overfit the dataset and the train and test set errors are comparable.\\
Further research could be done into whether the methods employed with the weighted sum in RNNs are actually able to identify semantically opposite sentences accurately. Also we could train embeddings and verify the performance on the rotten tomatoes dataset using RNNs. Training the entire dataset could be done to check the performance on the reverse dictionary application.
\bibliographystyle{IEEEtran}
\bibliography{one_word_bib}
\end{document}